# A novel filter based on three variables mutual information for dimensionality reduction and classification of hyperspectral images

[1] Asma Elmaizi*, [2] Elkebir Sarhrouni, [3] Ahmed hammouch, [4] Chafik Nacir

Laboratory LRGE, ENSET, Mohammed V University Rabat, Morocco

[1] asma.elmaizi@gmail.com, [2] sarhrouni436@yahoo.fr, [3] hammouch_a@yahoo.com, [4] nacir_chafik@yahoo.fr

*Abstract*— The high dimensionality of hyperspectral images (HSI) that contains more than hundred bands (images) for the same region called Ground Truth Map, often imposes a heavy computational burden for image processing and complicates the learning process. In fact, the removal of irrelevant, noisy and redundant bands helps increase the classification accuracy. Band selection filter based on "Mutual Information" is a common technique for dimensionality reduction.

In this paper, a categorization of dimensionality reduction methods according to the evaluation process is presented. Moreover, a new filter approach based on three variables mutual information is developed in order to measure band correlation for classification, it considers not only bands relevance but also bands interaction. The proposed approach is compared to a reproduced filter algorithm based on mutual information. Experimental results on HSI AVIRIS 92AV3C have shown that the proposed approach is very competitive, effective and outperforms the reproduced filter strategy performance.

I. INTRODUCTION

Since, for each pixel, hyperspectral sensors measure radiance values at a very large number of wavelengths, HSI contains a huge amount of data. Although the resolution provided allows material discrimination, the volume of data leads to many challenging problems such as data storage, computational efficiency and the "curse" of dimensionality [1]. The decrease of classification accuracy caused by redundant, and noisy bands that complicates the learning system and produces incorrect predictions, forced us to use the data reduction techniques to overcome these challenges.

This paper introduces a new approach to the HSI bands reduction and suggests that the proposed approach can provide a better HSI classification result by using filter bands subset selection. The approach was evaluated using HSI AVIRIS 92AV3C provided by the NASA [2].

The rest of the paper is organized as follows: part2 describes the dimensionality reduction techniques, while part3 reproduces a filter approach based on mutual information. Part4 presents the proposed method used to identify the most discriminative bands; part5 outlines the conducted experiment and presents its results. Finally, part6 concludes the paper.

II. CATEGORIZATION OF DIMENTIONNALITY REDUCTION METHODS ACORDING TO ATTRIBUTES EVALUATION PROCESS

Based on the generating attributes process [26], the dimensionality reduction can be done either by:

- Attributes Extraction where we transform the vectors of data.
- Attributes Selection without transformation of the data vectors.
- Selection followed by extraction of attributes.

According to the summary in [3], the dimensionality reduction methods generally include four basic steps see Figure 1:

1) *A procedure for generating candidate subsets by selection, extraction or extraction followed by selection.*
2) *An evaluation function to evaluate the current subset.*
3) *A stopping criterion for deciding when to stop the search.*
4) *A validation process to choose whether to keep the subset or not.*

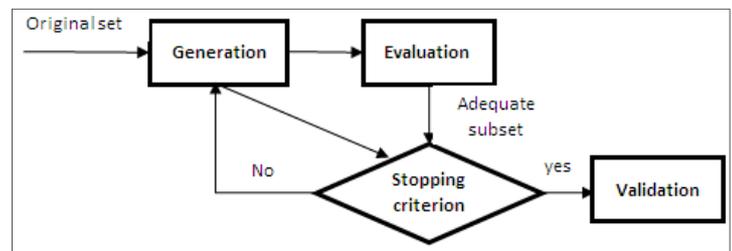

Fig 1: Dimensionality reduction Process with validation [3]

Hyperspectral band selection is one method of data reduction used in the pretreatment the classification step. The selection retains the physical meaning of the data set by selecting a set of bands from the input hyperspectral data set with high discriminative power leading to higher classification accuracy by discarding the irrelevant and redundant bands with the ground truth.

Performing bands selection helps improve [4]:

- Efficiency in terms of measurements costs, storage costs & computation costs.
- Classification performance.
- Ease of interpretation & Modeling.

Over the past few decades, many bands-selection approaches have been developed. they can be grouped into two categories:

wrapping and filter methods [3].

- The wrapping methods are classifier dependent. They use the prediction accuracy of the selected bands subset to measure its goodness and can be very computationally intensive. Classic approaches using this mode includes sequential forward selection (SFS) [19], sequential backward elimination (SBE) [19], and genetic algorithm (GA) [20]. Although the subset selected by these methods improves the classification performance, the wrapper-based method is computationally intensive because it has to run cross validation to avoid over fitting.

- The filter methods are classifier-independent; they select the subset attributes that maximizes a certain evaluation function. The lower ranked bands are removed, and the remaining bands are used to form a new subset as the input for classification. Representative methods of this category include correlation-based feature selection (CFS) [21], mutual information (MI) based selection [22][23]. These methods are easily applied to high-dimensional datasets, computationally simple and fast, and independent of the classification algorithm [24].

The filter methods are generally preferred to wrappers because of their usability with alternative classifiers, computational efficiency and simplicity. Therefore, the rest of the paper will focus on filter band selection
In literature, many researchers proposed different filter-selection methods using various measures as part of the evaluation function.
Information theory metrics based upon mutual information (MI) measure is one of the widely used measures in filtering methods. It evaluates the relative utility of each band to classification by measuring the amount of information that bands share with each other or with the ground truth [5].

## III. DIMENSIONALITY REDUCTION BY BAND SELECTION USING MUTUAL INFORMATION

The aim of this section is to reproduce a filter approach based on mutual information to reduce HSI dimensionality proposed by E.Sarhrouni [6].

### A. Definition and measure of Mutual information

In informative theory [7], mutual information (MI) measures the statistical dependence between two random variables.
To quantify the amount of information contained by a random variable A, Shannon entropy, denoted by H(A), is defined as:

$$H(A) = \sum_A p(A) \log_2 p(A) dA \quad (1)$$

Where p(A) is the probability density function (pdf) of A.

When two random variables A and B are given, the shared information between them can be measured by mutual information $I(A,B)$ as follows:

$$I(A, B) = \sum_{A,B} P(A, B) \log_2 \frac{P(A,B)}{P(A)p(B)} dAdB \quad (2)$$

Where p(A,B) denotes the joint pdf of A and B.

The relation between entropy and MI can be formulated as:

$$I(A, B) = H(A) + H(B) - H(AB) \quad (3)$$

When applied to band selection, the mutual information is used to find the relation between a band noted (A) and the ground truth noted (B). High Mutual Information indicates that the bands are strongly related, while zero Mutual Information shows that they are independent.
This relation is also called information gain; the band that has the highest mutual information with the ground truth is considered the most informative and discriminative one.

### B. Filter approach using mutual information

The basic idea of this filter approach proceeding by a forward selection is:

The band that has the largest value of mutual information with the ground truth is a good approximation of it. Thus, the subset of suitable bands is the one that generates the closest estimation to the ground truth (GT).
We generate the current estimation by the average of the last estimation of the GT with the candidate band.

The selection process "Algorithm1" is as follows:

**Algorithm 1 :** Extracted from [6]

1. *Order the bands according to decreasing value of their mutual information with the ground truth (GT).*

2. *Initialize the selected bands using the band that have the largest mutual information value with the GT.*

3. *Use the first selected band to build an approximation of the GT, denoted $GT_{est}$.*

4. *Calculate the mutual information between the GT and the GT => ($MI(GT, GT_{est})$.*

    *The last added band must increase the final value of $MI(GT_{est}, GT)$, otherwise, it will be rejected from the choices.*

5. *Introduce a threshold (Th) to control the permitted redundancy.*

### C. Discussion and Critics of the previous Method

The algorithm reproduced previously is based on feature selection using MI to select the bands that are able to classify the pixels of the Ground truth. Despite its simplicity and rapidity, it represents several limitations. The most drawbacks of the previous method are:

- Using a threshold that should be manually adjusted for controlling redundancy, this creates the risk that the algorithm may lose its selection action if we choose the wrong threshold. This will be further explained in the results section (V. B/C).

- According to bands evaluation process, maximizing the mutual information with the ground truth is not a very demanding criterion. Since the algorithm will eliminate irrelevant bands based on their correlation with the GT, without thinking that, they can be very important for discriminating the ground truth when combined with other bands. The previous approach neglected the complementarity between the selected bands.

## IV. THE PROPOSED APPROACH

### A. Definition and measurement of the interaction information

Within the informative theory researches, the idea of employing three-variable mutual information or interaction gain $I(A,B,C)$ have been proposed by Jakulin [8][11] in order to study the relation between the feature(A) and the class(C) in the context of the other feature(B).

Interaction information is defined [8] as the decrease in uncertainty caused by joining attributes A and B in a Cartesian product.

$$I(A,B,C)=I(A;B,C)-I(A,C)-I(B,C) \qquad (4)$$

The value of the interaction is the same, even when the class label is swapped with one of the features in Equation (4).

### B. The proposed algorithm

In this algorithm2, we will subjoin to the mutual information used previously in algorithm1, the three-variable mutual information defined in part A in order to ensure maximum relevance and maximum interaction of the selected bands.

We generate the current estimation by the average of the last estimation of the GT with the candidate band.

Our process of band selection "Algorithm2" is as follows:

**Algorithm 2 :**

1. *Order the bands according to decreasing value of their mutual information with the ground truth (GT).*

2. *Initialize the selected bands by the band that have the largest mutual information value with the GT.*

3. *Use the first selected band to build an approximation of the GT, denoted $GT_{est}$.*

4. *Calculate the three variables mutual information between the candidate band (B), the GT and $GT_{est}$ => $I3 (GT_{est}, B, GT)$, The $GT_{est}$ is calculated by averaging the last estimation of the GT with*

5. *Add the relevant band that maximizes I3 calculated previously.*

## V. EXPERIMENT, RESULTS AND DISCUSSION

### A. Case study & SVM classifier

To test the performance of the methods, the experiment was conducted on the Hyperspectral image AVIRIS 92AV3C [2]. The Indian Pines image was gathered by Airborne Visible/Infrared Imaging Spectrometer (AVIRIS) sensor over Northwest Indiana's Indian Pines test site in June 1992. This image consists of $145 \times 145$ pixels and has the spatial resolution of 20 m/pixel. The AVIRIS sensor generates 220 spectral bands in the wavelength range of 0.2−2.4 µm illustrates in the 3d cube figure 2.

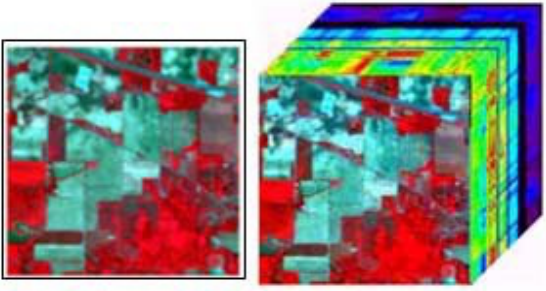

Fig 2. 3D cube of Indiana pines data set (in the right), three-band color composite image (in the left)

The ground truth map is also provided, but only 10366 pixels are labeled. Each label indicates one from 16 classes introduced in figure 3. 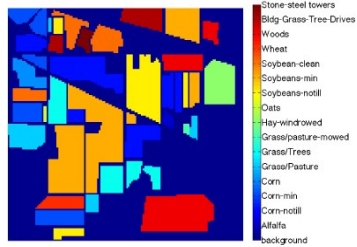

Fig 3. The Ground Truth map of AVIRIS 92AV3C

50% labeled pixels are randomly selected to be used in training, and the other 50% will be used for the classification testing [5]. The classifier used is the support vector machine ''SVM'' [9] [10].
The average classification accuracy is used as a performance measure.

### A. Results

The table 1 gives the classification results using the selection by mutual information (process1) for different thresholds Th using the SVM classifier [6].

The following table 2 presents the results obtained using the proposed method algorithm 2 based on three variables mutual information and the classifier SVM.
Results demonstrates the effectiveness selection of this algorithm.

| | The accuracy (%) of classification for numerous thresholds | | | |
|---|---|---|---|---|
| Th | -0,02 | -0,01 | -0,0040 | -0,000 |
| 2 | 47,44 | 47,44 | 47,44 | 47,44 |
| 3 | 47,87 | 47,87 | 47,87 | 48,92 |
| 4 | 49,31 | 49,31 | 49,31 | |
| 12 | 56,30 | 56,30 | 56,30 | |
| 14 | 57,00 | 57,00 | 57,00 | |
| 18 | 59,09 | 59,09 | 62,61 | |
| 20 | 63,08 | 63,08 | 63,55 | |
| 25 | 66,12 | 64,89 | 65,38 | |
| 35 | 76,06 | 74,72 | | |
| 36 | 76,49 | 76,60 | | |
| 40 | 78,96 | 79,29 | | |
| 45 | 80,85 | 81,01 | | |
| 50 | 81,63 | 81,12 | | |
| 53 | 82,27 | 86,03 | | |
| 60 | 82,74 | 85,08 | | |
| 70 | 86,95 | | | |
| 75 | 86,81 | | | |
| 80 | 87,28 | | | |
| 83 | 88,14 | | | |

(Number of retained Bands in leftmost column)

Tab 1. Results of algorithm 1 [6]: redundancy control for different values of threshold (Th)

| Number of retained Bands | The accuracy (%) of classification |
|---|---|
| 3 | 55,99 |
| 4 | 59,36 |
| 12 | 69,32 |
| 14 | 78,30 |
| 18 | 80,05 |
| 20 | 80,67 |
| 25 | 82,02 |
| 35 | 82,82 |
| 36 | 82,97 |
| 40 | 83,73 |
| 45 | 84,51 |
| 50 | 85,45 |
| 53 | 85,53 |
| 60 | 86,30 |
| 70 | 88,17 |
| 75 | 88,02 |
| 80 | 88,25 |
| 83 | 88,19 |

Tab 2. Results of algorithm 2: redundancy and interaction control

Figure 4 shows the average classification on the AVIRIS 92AV3C [2], it is calculated over the hole size of the selected subset, starting from 2 bands up to 83 bands.

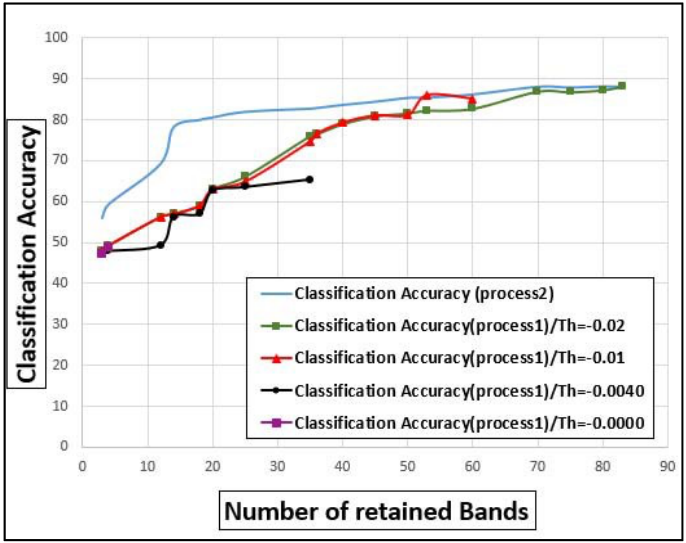

Fig 4. Average classification accuracy comparison results between both algorithms

We illustrate in Figure .5, the Ground Truth map originally displayed in figure 3, and the scene classified with our method, for 20 selected bands.

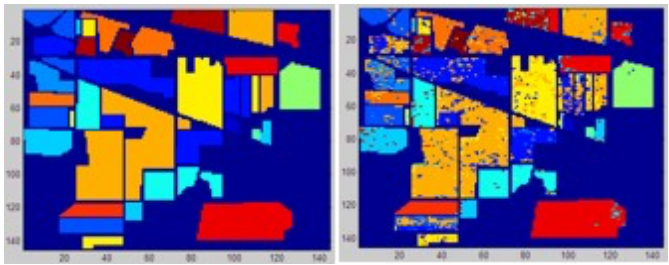

Fig 5. Original Grand Truth map(in the left) and the map produced by our approach according to 20 bands (in the right). Acc*uracy=80,67%.*

### B. Analysis & discussion

The above graph (figure 4) and the figure 5 shows the strength of our proposed approach compared with mutual information selection filter. The analysis of the graph and the previous tables allowed us to take out the following results:
According to table 1, we can see that:

- For high thresholds values, in the range of (-0.004 to 0), few bands are selected because there is no redundancy, for example, with Th=0, just three bands are retained.
If we allowed some redundancy, for thresholds (-0.02 to -0.005), the classification rate increases and achieves 88% with Th=-0.02 and 83 selected bands.

- The threshold should be adjusted correctly, so the algorithm 1 would not lose its selection action.

According to table 2 and the graph in figure 4, we notice that:

- As shown, the proposed algorithm achieves the highest accuracy and outperforms the first algorithm. The proposed approach selects bands with high discriminative power very quickly. It achieves 80% classification accuracy with 20 bands, which is higher than algorithm 1 (Th=-0,02) by 17, 60%.

- we can note that 20 bands are sufficient to detect materials contained in the region as shown in figure 5.

- With 70 bands, the proposed approach produced its best accuracy of 88%, which is better than algorithm 1 (Th=-0,02) about 1, 22%.

- For the number of retained bands upper than 83, the algorithm reaches its maximum accuracy value =88%. The added bands do not contain any relevant information for discrimination. They are either redundant with the already selected ones or noisy witch saturate the classification accuracy at 88%.

- In comparison with the wrapper approach based on mutual information and the error probability proposed by E.sarhrouni [25], our filter outperforms this wrapper for values 20 to 80 retained bands (17% for 20 bands). For values upper than 80 bands the wrapper outperforms our filter by 2%.

- We can say that our filter works faster and outperforms the wrapper results on a certain range by using a simple measurement of I3. However, its result is not always satisfactory (upper than 80 retained bands). On the other hand, the wrapper guarantees good results through examining learning results, but it is very slow.

- We can conclude that the proposed algorithm uses a simple evaluation function based on the three variables mutual information without including a threshold that could limit the selection process as shown in figure 4. Its major advantage is the redundancy and interaction control. At the same time, the proposed approach is slower than the algorithm [6] due to the computation of I3 that is more complex than MI.

## VI. CONCLUSION

This paper presents a new bands-selection filter method based on three variables mutual information to solve the problem of bands interaction in Hyperspectral images in order to reduce their dimensionality. The proposed filter selects relevant and non-redundant bands by taking into consideration the correlation between the candidate bands and the bands already selected within the subset. The method is compared with a reproduced filter approach based on mutual information by conducting an experiment on HSI AVIRIS 92AV3C using the support vector machine classifier. The results shows that the proposed algorithm improves the classification accuracy with a lesser number of bands compared to the MI filter-based approach. The results demonstrate the ability of the proposed method to speedily select a bands subset with high discriminative power and confirm that the analysis of the interaction and correlation between bands must be taken into consideration for speedily selecting bands with high discriminative ability for classification task.

The proposed method is an improved filter that can be used in fast applications and that will be investigated and developed in the future for a better discrimination, reduction and classification of hyperspectral images.